\documentclass[runningheads]{llncs}
\usepackage[]{graphicx}
\usepackage{url}
\usepackage{subcaption}
\usepackage{array}
\usepackage{placeins}
\usepackage{amssymb,amsmath}

\usepackage[usenames, dvipsnames]{color}

\title{Rotational 3D Texture Classification Using Group Equivariant CNNs}

\author{Vincent Andrearczyk  $^{1}$
\and Adrien Depeursinge $^{1,2}$}
\authorrunning{Vincent Andrearczyk and Adrien Depeursinge}
\institute{$^1$Institute of Information Systems, University of Applied Sciences Western Switzerland (HES-SO), Sierre, Switzerland \\
 $^2$Nuclear Medicine and Molecular Imaging Department, Lausanne University Hospital, 
Lausanne, Switzerland}
\toctitle{Lecture Notes in Computer Science}
\tocauthor{Authors' Instructions}

\begin{document} 
\maketitle 
\begin{abstract}
Convolutional Neural Networks (CNNs) traditionally encode translation equivariance via the convolution operation. Generalization to other transformations has recently received attraction to encode the knowledge of the data geometry in group convolution operations. Equivariance to rotation is particularly important for 3D image analysis due to the large diversity of possible pattern orientations. 3D texture is a particularly important cue for the analysis of medical images such as CT and MRI scans as it describes different types of tissues and lesions. In this paper, we evaluate the use of 3D group equivariant CNNs accounting for the simplified group of right-angle rotations to classify 3D synthetic textures from a publicly available dataset. The results validate the importance of rotation equivariance in a controlled setup and yet motivate the use of a finer coverage of orientations in order to obtain equivariance to realistic rotations present in 3D textures.
\end{abstract}
%
%
%
\section{Introduction}
\label{sec:intro}
CNNs naturally benefit from the translation equivariance of the convolution operation. They have been used in various studies to analyze static or dynamic textures by incorporating an orderless pooling of feature maps to discard the overall shape and layout analysis and, thus, describe repetitive texture patterns \cite{AnW2016,CMK2016,ZXD2016}. 
Robustness to other groups of transformations, including rotation and reflection, have been traditionally approximated by showing various examples of transformed data, e.g. in the form of naturally large training sets or artificial data augmentation.
The cost of learning such transformations has been mitigated by transfer learning, i.e. re-using weights pre-trained on a very large dataset, typically ImageNet~\cite{RDS2015} for 2D images. The features learned on such large datasets (e.g. early colors, edges and more complex shapes) have proven to generalize well to various computer vision tasks \cite{AnW2016,CMK2016}. 
However, such models also require larger Degrees of Freedom (DoF), and the type of invariance required to perform a given task may depend on the data and task. 
For instance, analyzing local patterns (e.g. tumor or organ walls, diverse vascular configurations) in medical images often require local rotation invariance. 
On the other hand, the scale is generally controlled and can become an informative feature that we do not want to discard by invariance.
Besides, the amount of data available in medical imaging is often limited and cannot include all possible 3D orientations of all patterns relevant to the analysis task at hand.
It follows that controlling the type of desired equivariances in a CNN is an important characteristic 
in terms of speed of convergence and data efficiency. 

Hard-coding equivariance into CNNs has gained a strong interest in recent literature \cite{CoW2016b,cohen2016steerable,dumont2018robustness,weiler2017learning,winkels20183d,worrall2018cubenet,weiler20183d}.
Group equivariant convolutions~\cite{CoW2016b} ($G$-convolutions) exploit weight sharing across transformation channels, in particular right-angle rotations and roto-reflections, expanding the expressive capacity of the network with a limited increase in the number of parameters.
The use of 2D steerability was also proposed~\cite{weiler2017learning,WGT2016} for advanced rotation equivariance.

3D CNNs~\cite{ji20133d} will largely benefit from built-in rotation equivariance due to the larger degree of possible 3D rotations (i.e. three rotation axes) that must be otherwise learned via data augmentation and unnecessary additional DoFs.
3D $G$-CNNs were recently developed for pulmonary nodule detection~\cite{winkels20183d} in chest CT scans and for 3D object recognition and volumetric boundary segmentation~\cite{worrall2018cubenet}.

In this paper, we evaluate the importance of equivariance and invariance to rotations for classifying 3D textures using classic and group equivariant CNN designs. We employ a controlled experimental setup with synthetic texture volumes having similar characteristics as organ and tumor tissue in medical imaging.

\section{Methods}
\label{sec:methods}

\subsection{Group Equivariant CNN}
\label{sec:gcnn}
A convolution layer uses the translation symmetry of grid data by sharing weights across spatial shifts, resulting in a translation equivariance ($f(T\boldsymbol{x})=Tf(\boldsymbol{x})$) maintained throughout the network.
This equivariance is generally pooled into a translation invariance by the last pooling and fully connected layers ($f(T\boldsymbol{x})=f(\boldsymbol{x})$).
Here, $f:\mathbb{Z}^n\rightarrow\mathbb{R}^K$ acts on a stack of $K$ $n$-dimensional grids, i.e. $n$-D image or feature maps  ($n=3$ in this paper). 
The transformation operator $T$ implements, in this case, an $n$-D translation.
Yet, the weight sharing can be extended to larger transformation groups $G$ to exploit the symmetries of the data, including rotations and reflections. 
Standard convolutions and their group equivariant counterparts ($G$-convolutions) were formally defined in Cohen and Welling~\cite{CoW2016b}. 
A $G$-convolution is equivariant to the translation group and to a symmetry group $H$, together forming the larger group $G$. In this paper, $H$ is the cube rotation group $O$ or roto-reflection $O_h$\footnote{The lowerscript $h$ here is not related to a transformation $h\in H$.}. 
The $G$-convolution output contains a number of feature maps $K=|H|\times M$, where $|H|$ is the number of elements in the group $H$, and $M$ is the number of filters. The $M$ filters are typically transformed by all elements in $H$ and the result of their convolution with the input channels are organized as orientation channels.
With this built-in equivariance, we have $f(h\boldsymbol{x})=hf(\boldsymbol{x})$, i.e. a transform of the input by $h\in H$ results in the same transform of the feature maps. 
The orientation channels of the filters, however, undergo a permutation to maintain the equivariance throughout the network as described in~\cite{winkels20183d}.
$G$-pooling allows pooling the activations over the $|H|$ transformation channels to obtain local or global invariance to the symmetry group $H$. 
It was shown in \cite{CoW2016b} that premature invariance in early layers was not desirable in their experiment, yet Local Rotation Invariance (LRI) can be useful for texture analysis as discussed in Section \ref{sec:news}.

In the 3D case~\cite{winkels20183d}, we consider equivariance to orientation-preserving rotations of the cube, i.e. the $O$ group including $|O|=24$ elements, as well as the full symmetry group of the cube (roto-reflection), i.e. the $O_{h}$ group with $|O_h|=48$ elements. 
Note that in medical imaging, the pixel spacing in the $z$-direction may differ from the planar one, e.g. in CT and MRI volumes. In this case, one would consider rectangular cuboid rotations ($D_4$ and $D_{4_h}$).
%
\subsection{Dataset} 
\label{sec:dataset}
\begin{figure}[ht]
\centering
\begin{subfigure}{.35\textwidth}
  \centering
  \includegraphics[width=.7\linewidth,trim={0.5cm 0.5cm 0.5cm 0.5cm}]{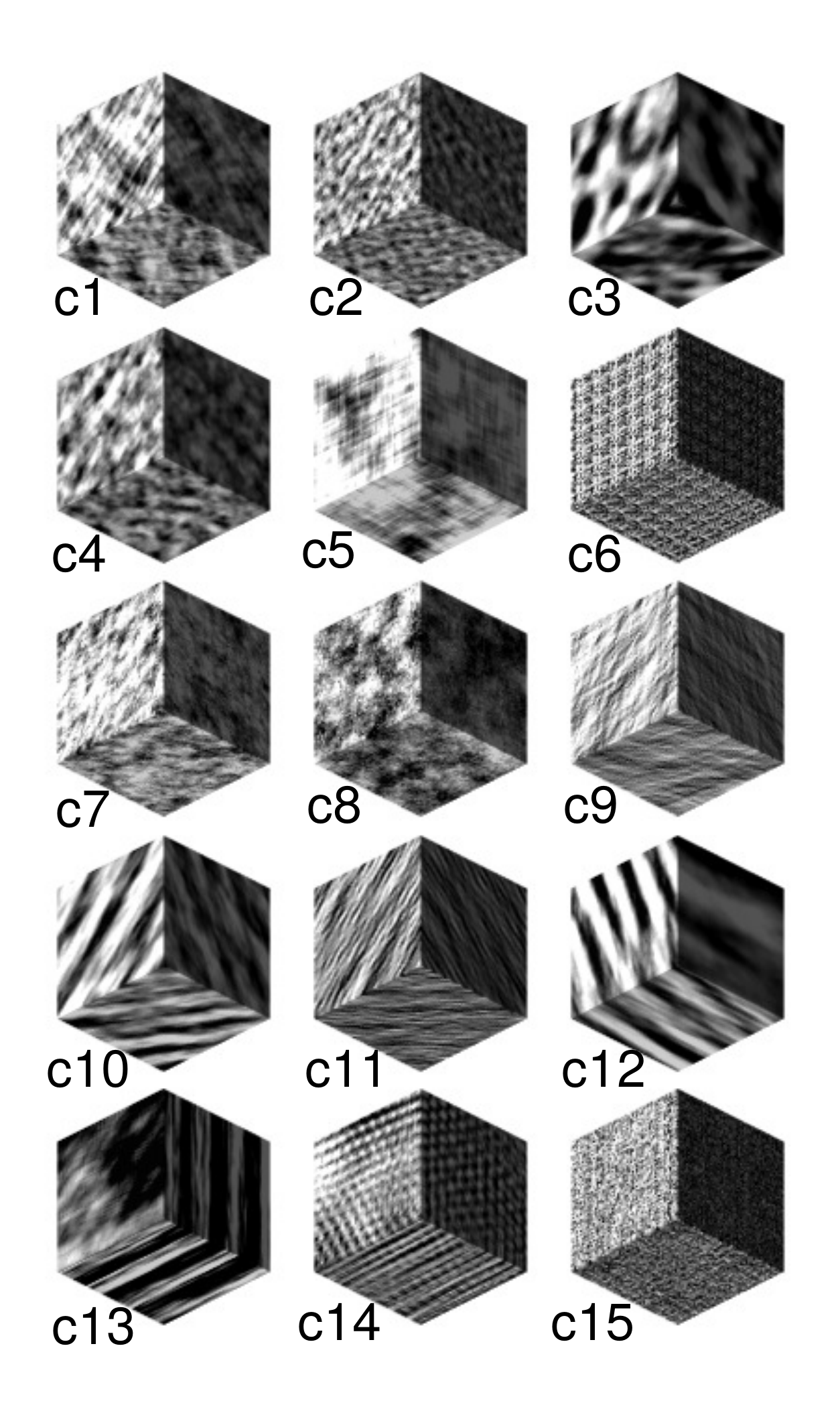}
\end{subfigure}
\begin{subfigure}{.6\textwidth}
  \centering
  \includegraphics[width=.7\linewidth,trim={0.5cm 0.5cm 0.5cm 0.5cm}]{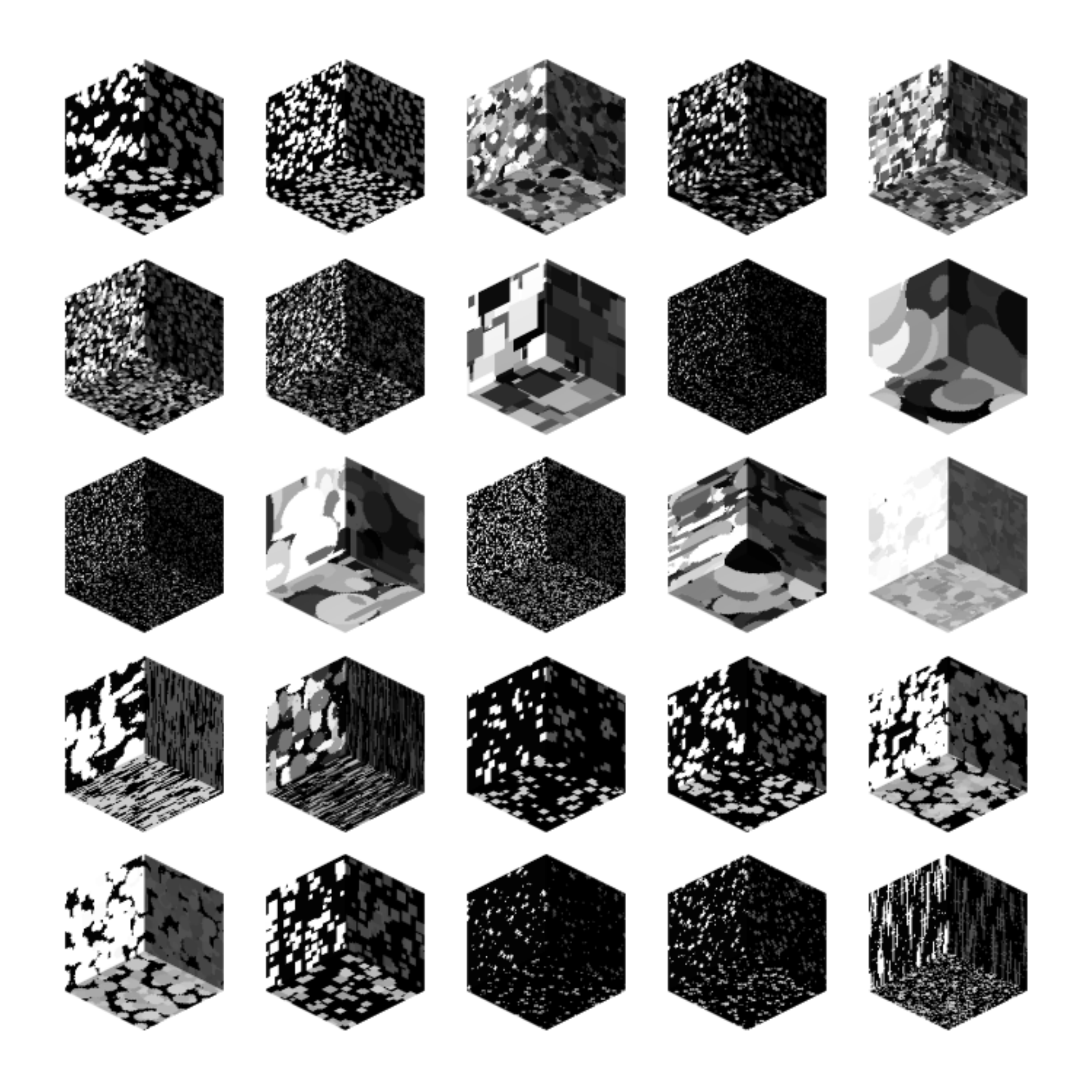}
\end{subfigure}
\caption{The Fourier (left) and Geometric (right) RFAI datasets are composed of 15 and 25 classes respectively. Each instance is a 3D image of $64\times 64\times 64$ voxels.
}
\label{fig:samples}
\end{figure}

Various medical image modalities are characterized by 3D textures such as MRI and CT scans of most organs including the lung, liver and breast. 
In this paper, we focus on synthetic data to obtain a controlled setup with representative texture classes and known transformations in the training and test data.
The RFAI (Reconnaissance de Formes, Analyse d'Images) database \cite{paulhac2009solid} contains 3D synthetic texture volumes having similar characteristics as organ and tumor tissue in medical imaging.
We evaluate the methods on the Fourier and Geometric datasets of the RFAI database, composed of 15 and 25 classes respectively and 10 instances per class. 
The limited amount of data is representative of many medical imaging tasks due to patient privacy and to the cost of data annotation.
The Geometric dataset was constructed using random positioning of geometric shapes such as spheres, cubes, and ellipses, while the Fourier dataset exhibits more directional textures and was constructed from synthetic distributions in the Fourier domain.
Each instance is a 3D image of $64\times 64\times 64$ voxels. 
Fig. \ref{fig:samples} illustrates one volume instance of each class.

For each instance of the datasets, 
a rotated version is provided~\cite{paulhac2009solid} to evaluate and compare the rotation invariance of the algorithms.
Finally, we introduce another rotated test set that we refer to as $O$-rotated as the rotations applied to the texture volumes are randomly chosen from the group of cube symmetries $O$ (i.e. right-angle rotations). We keep the same random rotations for all experiments for a fair comparison.
The data is normalized with zero mean and unit variance. 

We evaluate the method with a k-fold Cross-Validation (CV), where $k=15$ for the Fourier dataset and $k=25$ for the Geometric one. The random splits are kept unchanged for all experiments.
Results based on a leave-one-out (LOO) CV as reported in \cite{depeursinge2018rotation,paulhac2009solid} should be similar or better than the ones reported in this paper as the LOO CV uses more training data for each fold. 
For each run, we train using the normal set ($k-1$ folds) and test either on the normal, $O$-rotate or rotate sets for the remaining fold.
Data augmentation is not performed since we want to evaluate the contribution of group equivariance in the context of a small amount of controlled training data.
\subsection{Network Architectures}
\label{sec:net_arch}
\begin{figure}[ht]
	\centering
	\includegraphics[width=0.7\textwidth]{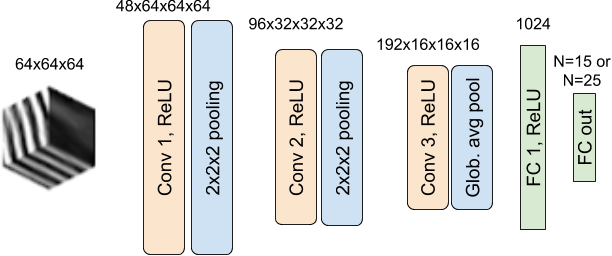}
	\caption{The $Z3$-CNN architecture. For the $G$-CNNs, the architecture is similar but with a number of filters divided by $\sqrt{|H|}$ and the feature maps are extended to orientation channels.}
	\label{fig:architecture}
\end{figure}

As a baseline, we used a standard CNN without group equivariance that we refer to as $Z3$-CNN, summarized in Fig.~\ref{fig:architecture}.
The $Z3$-CNN is composed of three $3\times3\times3$ convolution layers with 48, 96 and 192 output feature maps respectively. The first and second convolutions are down-sampled by $2\times2\times2$ max-pooling operations. 
We average the feature maps across the spatial locations (global average pooling) after the third convolution layer 
in order to learn and densely pool repeated texture features~\cite{AnW2016}. 
A first dense layer (i.e. fully connected, FC) with 1024 neurons is used, as well as an output dense layer with $N$ neurons ($N$ being the number of classes).
The convolutional and dense layers are activated by a ReLU, except for the output dense layer activated by a softmax function.

We then derive several group equivariant networks from the $Z3$-CNN baseline. 
After the global average pooling, we can directly connect the orientation selective features maps to the fully connected layers ($G^{fc}$-CNN). 
We can also use maximum or average $G$-pooling ($G^{max}$-CNN and $G^{avg}$-CNN) to pool across the orientations.
We do not incorporate $G$-pooling in early layers as it has been shown to reduce performance~\cite{CoW2016b}.
In order to maintain the number of free parameters roughly equal across networks (except for the $G^{fc}$-CNNs as reported in Table~\ref{tab:res}), we divide the number of filters in the convolution layers by $\sqrt{|H|}$, with $|O|=24$ and $|O_h|=48$.

The weights are initialized with the Xavier method~\cite{glorot2010understanding}.
An Adam optimizer is used with standard hyperparameters ($\beta_1=0.9$, $\beta_2=0.999$ and $\epsilon=10^{-8}$) to optimize the cross-entropy loss. The initial learning rate is set to $10^{-4}$ and a batch size of 16 is used.
\section{Experimental Results}
\label{sec:results}

The results are summarized in Table \ref{tab:res}.
For such a small dataset, shallow hand-crafted features tailored to rotation invariance \cite{depeursinge2018rotation} naturally perform better than the complex 3D CNNs with only equivariance to rotations in $O$ and $O_h$. 
The goal of these experiments is not to compare against these methods as a 3D CNN with pre-trained weights and data augmentation would easily obtain around $100\%$ accuracy. The point is rather to evaluate the built-in equivariance and invariance of these networks.
The confusion matrices of the $Z3$-CNN and $O$-CNN on the Fourier $O$-rotate are reported in Fig.~\ref{fig:samples}.

\begin{table}[h]
\centering
\caption{Results of the k-fold overall accuracy and standard deviation ($\%$) on the Fourier (F.) and Geometric (G.) RFAI datasets. Chance level accuracy is 6.7$\%$ and 4$\%$, respectively. The test volumes are taken from either the normal (norm.), the $O$-rotate ($O$-rot.) or the rotate (rot.) datasets.}
\begin{tabular}{lccc|ccc}	
  	Network (\# param.)			& F. norm. 	& F. $O$-rot. & F. rot. 	& G. norm. 	& G. $O$-rot. & G. rot. 	\\
  	\hline  			
  	  $Z3$-CNN (271,039)		& 94.0\tiny{$\pm{6.11}$}		& 70.0\tiny{$\pm{14.1}$}		& 62.7\tiny{$\pm{18.1}$} 		& 88.8\tiny{$\pm{9.9}$}		& 86.4\tiny{$\pm{11.6}$}		& 77.6\tiny{$\pm{11.8}$} 		\\
  	  $O^{fc}$-CNN (670,054) & 95.3\tiny{$\pm{7.2}$}		& 49.3\tiny{$\pm{22.3}$}		& 54.0\tiny{$\pm{19.9}$} 		& $\boldsymbol{97.6}$\tiny{$\pm{5.8}$}			& 89.6\tiny{$\pm{11.8}$}			& 78.8\tiny{$\pm{12.4}$} 		\\
  	  $O^{max}$-CNN (199,014)& 96.0\tiny{$\pm{4.9}$}		& 95.3\tiny{$\pm{6.2}$}		& $\boldsymbol{77.3}$\tiny{$\pm{12.9}$} 		& 96.0\tiny{$\pm{5.6}$}			& $\boldsymbol{97.6}$\tiny{$\pm{5.1}$}			& 87.6\tiny{$\pm{10.7}$} 		\\
  	  $O^{avg}$-CNN (199,014) & 92.7\tiny{$\pm{8.5}$}		& 92.0\tiny{$\pm{7.5}$}			& 70.7\tiny{$\pm{12.4}$} 		& 96.0\tiny{$\pm{4.8}$}			& 96.8\tiny{$\pm{5.4}$}			& $\boldsymbol{88.8}$\tiny{$\pm{10.7}$} 		\\ 
  	  $O_h^{fc}$-CNN (1,020,549)& 94.7\tiny{$\pm{6.2}$}			& 52.7\tiny{$\pm{18.8}$} 		& 48.7\tiny{$\pm{20.9}$}			& 95.2\tiny{$\pm{12.6}$}			& 88.3\tiny{$\pm{13.7}$}			& 81.2\tiny{$\pm{4.0}$} 		\\
  	  $O_h^{max}$-CNN (250,501)& $\boldsymbol{96.7}$\tiny{$\pm{6.0}$}			& $\boldsymbol{96.7}$\tiny{$\pm{6.0}$}			& 75.4\tiny{$\pm{16.7}$} 		& 95.2\tiny{$\pm{10.9}$}			& 96.8\tiny{$\pm{12.9}$}			& 88.3\tiny{$\pm{7.2}$} 		\\
  	  $O_h^{avg}$-CNN (250,501)& 89.7\tiny{$\pm{12.5}$}			& 88.8\tiny{$\pm{11.1}$}			& 70.2\tiny{$\pm{11.9}$} 		& 92.4\tiny{$\pm{15.5}$}			& 92.0\tiny{$\pm{14.9}$}			& 86.4\tiny{$\pm{9.4}$} 		\\
  	\hline 
\end{tabular}
\label{tab:res}
\end{table}

%

\begin{figure}[ht]
\centering
\begin{subfigure}{.495\textwidth}
  \centering
  \includegraphics[width=1.05\linewidth,trim={0.8cm 0 0.85cm 0}]{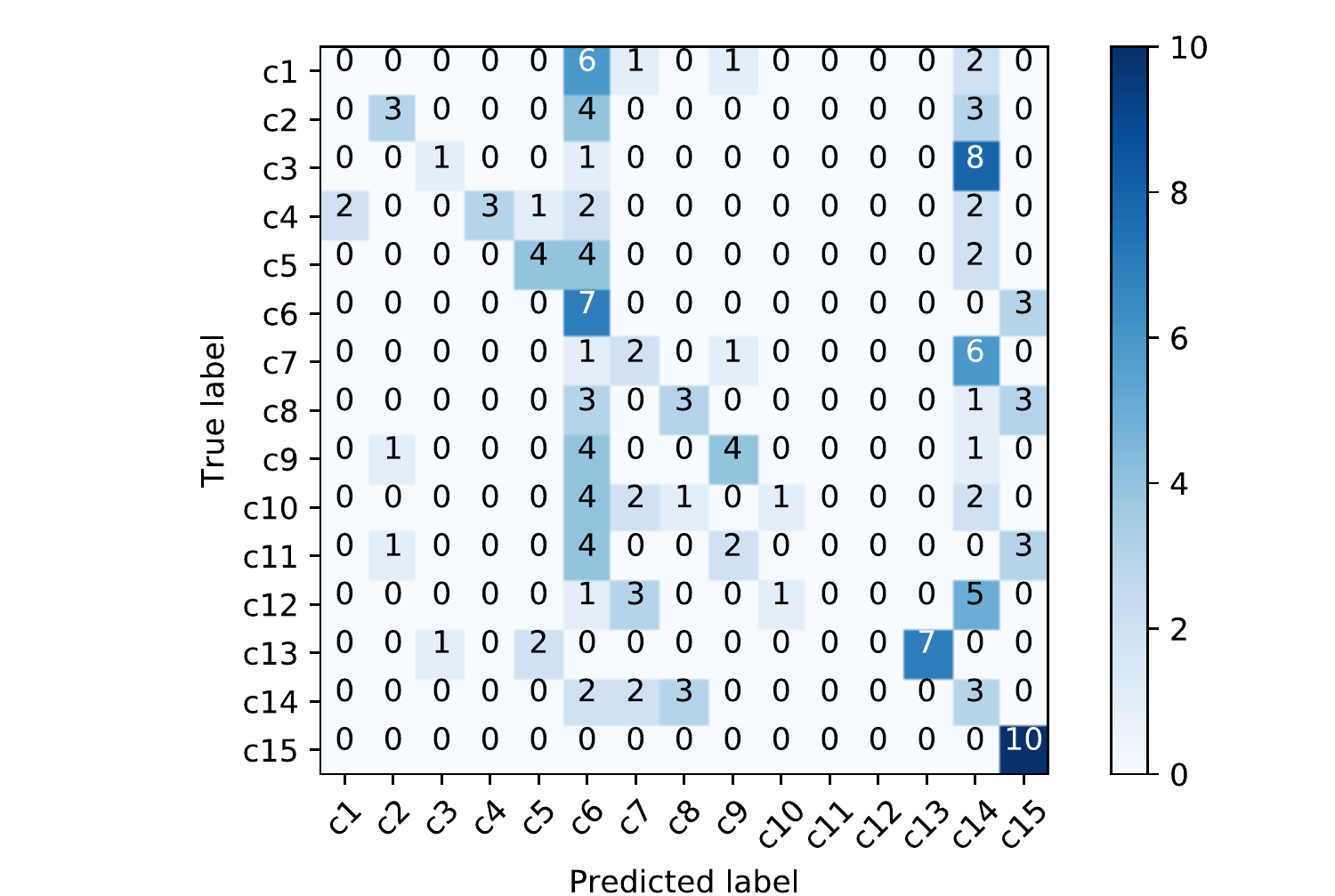}
\end{subfigure}
\begin{subfigure}{.495\textwidth}
  \centering
  \includegraphics[width=1.05\linewidth,trim={0.8cm 0 0.85cm 0}]{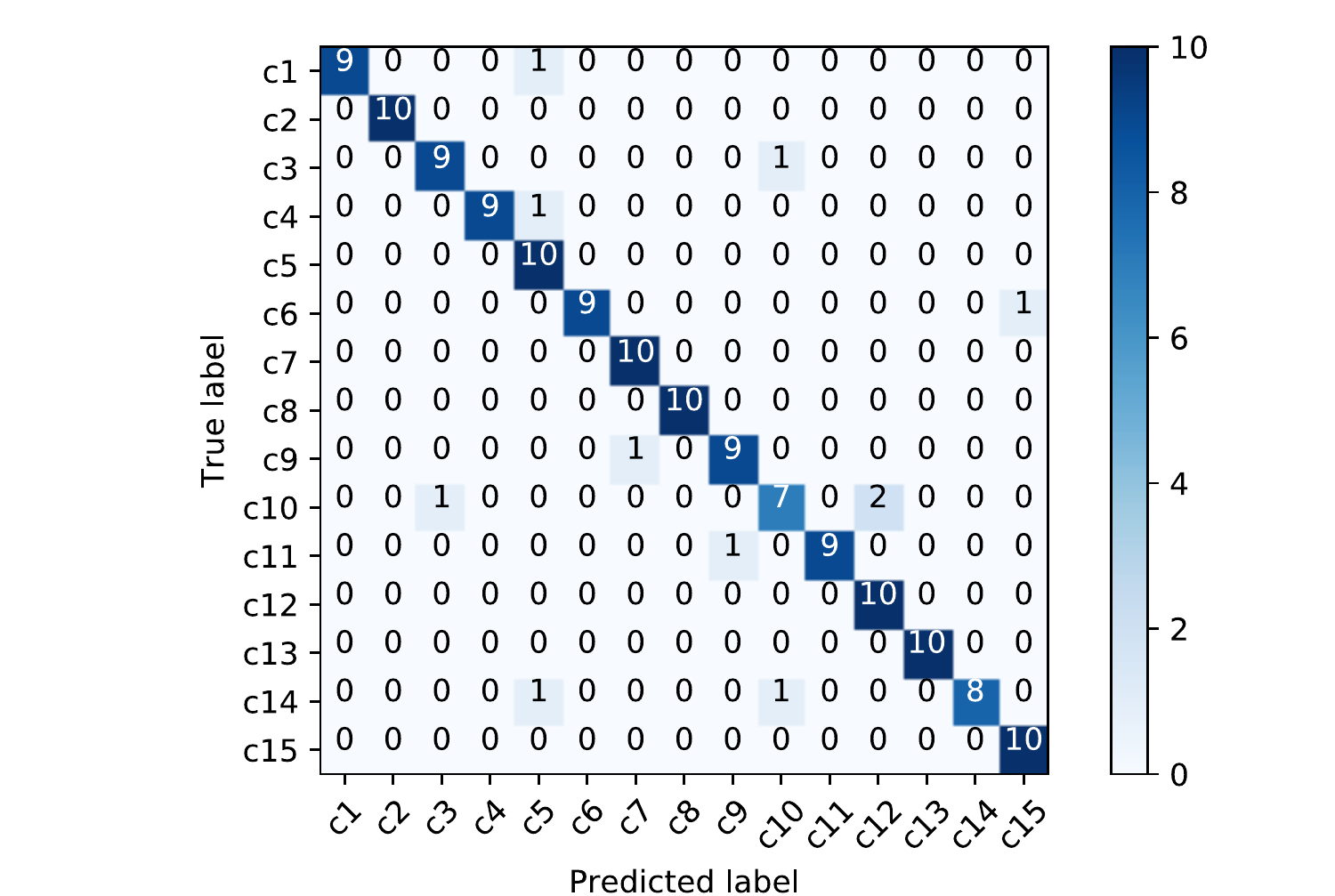}
\end{subfigure}
\caption{Confusion matrices on the Fourier $O$-rotate set with the $Z3$-CNN (left) and $O^{max}$-CNN right).}
\label{fig:cm}
\end{figure}

\section{Discussion}
\label{sec:news}
On the normal set, best results are obtained with the $G$-CNNs using either no $G$-pooling or max $G$-pooling.
It is worth noting that the $G$-CNNs have fewer DoFs to model the diversity of texture patterns sought by the network.
This is due to maintaining approximately the same number of free parameters by dividing by $\sqrt{|H|}$. In this way, we enforce the rotation equivariance by weight sharing and reduce the DoFs in the patterns learning.
The rotation equivariance being unnecessary for the normal test set, it is remarkable that such high performance is obtained by the $G$-CNNs, showcasing its ability to leverage data symmetries by sparing the parameter budget on data diversity.

The $Z3$-CNN does not generalize well to the rotated datasets, yet it performs better on the $O$-rotate set than the rotate one. 
In particular, the Geometric dataset containing various simple shapes such as cubes, the right-angle rotations result in less variation from the training set than the random rotations.
The results on the $O$-rotate set show that invariance to right-angle 3D rotations is obtained by the $O$-CNN with a max or average $G$-pooling, although the latter reduces the discriminability by averaging across the orientations.
The $O$-CNN without $G$-pooling is not invariant and performs only slightly better than the $Z3$-CNN on the $O$-rotate set.
These results show that, with limited training data, equivariance is not sufficient and the invariance must be encoded with $G$-pooling.

On the rotate set, the $G$-CNNs outperform the $Z3$-CNN, yet a significant drop of accuracy is observed as compared to the $O$-rotate set, suggesting that the right-angle symmetries do not extract information in enough orientations to sufficiently describe the possible patterns orientations. This result motivates the further use of higher order rotational symmetries and steerable filters as previously proposed in 2D~\cite{weiler2017learning} and recently in 3D~\cite{weiler20183d}. 
The poor generalization of the $O$-CNN on the rotate set can also be explained by the lack of variability in the directionality of the training textures (all volumes of a class have the same directionality in the normal set), resulting in very directionally selective filters. This precision is useful for the normal set, yet any small rotation of the textures results in a mismatch with the convolution filters. 
Based on the analysis of confusion matrices, we observed that textures with directionality benefit more from the rotation invariance than non-directional textures. 
To illustrate this, we report the confusion matrices of the $Z3$-CNN and $O$-CNN on the Fourier $O$-rotate set and compare the accuracy on the different texture classes shown in Fig.~\ref{fig:samples}. 
Volume examples of the classes are shown in Fig.~\ref{fig:samples} with the corresponding class numbers (c1 to c15). Non-directional texture classes such as c6 and c15 are already correctly classified by the $Z3$-CNN, while highly directional textures such as c10, c11 and c12, benefit greatly from the invariance. 
The directionality of the textures of class c13 is aligned with two rotation axes (see Fig.~\ref{fig:samples}). Therefore, the $Z3$-CNN is less affected by right-angle rotations of the volumes as they are likely to result in the same texture directionality (7 out of 10 correctly classified).

The $O_h$ equivariance does not result in a significant increase of accuracy as compared to the $O$-CNN. 
This is due to the $O$-rotate and rotate sets being obtained with orientation preserving rotations. For this task, the enforced equivariance to reflections is unnecessary and reduces further the number of filter patterns that the network can learn.

Combining Directional Sensitivity (DS) and LRI is an important aspect of 3D texture analysis in real medical data \cite{depeursinge2018rotation}.
With the $G$-CNNs, DS is learned throughout training which is made easier with the weight sharing across filters orientations.
The LRI can be obtained by performing the $G$-pooling before global average pooling, with a local support equivalent to the effective receptive field of the neurons at the orientation pooling layer.
We are currently investigating the LRI for right-angle rotations at different scales using $G$-pooling at various depths.
On the Fourier and Geometric datasets, the LRI before global average pooling for right-angle rotations with max $G$-pooling results in a small drop of accuracy (approximately 2\% drop) on the normal and $O$-rotate sets.
%
\section{Conclusion}
\label{sec:conclusions}
In this paper, we evaluated the use of $G$-CNNs for the classification of 3D synthetic textures representative of rotated texture patterns in medical imaging.
We experimentally confirmed the importance of using CNN designs including built-in equivariance and invariance (when using $G$-pooling) to the $O$ group, which can be easily extended to other symmetry groups. The results showed that on this simple dataset, the equivariance to right-angle rotations is very useful but yet not sufficient to analyze randomly rotated textures. This observation, therefore, motivates the use of a higher order rotational symmetry, as proposed in 2D CNNs with steerable filters~\cite{weiler2017learning,WGT2016} and recently in 3D~\cite{weiler20183d}. 
In 3D, the model size, the computational complexity of 3D filters and of feature maps and the number of symmetries $|H|$ to encode are significantly larger than in 2D, resulting in a considerable bottleneck in terms of GPU resources.
%
\paragraph{Acknowledgements}
This work was supported by the Swiss National Science Foundation (grants PZ00P2\_154891 and 205320\_179069).
\bibliographystyle{spiebib} 
\bibliography{mybibfile}
\end{document}